# Selection of future events from a time series in relation to estimations of forecasting uncertainty

Igor B. Konovalov

*Institute of Applied Physics, Russian Academy of Sciences, 46 Ulyanov Str., Nizhny Novgorod, 603950 Russia E-mail: konov@appl.sci-nnov.ru*

**Abstract:** A new general procedure for a priori selection of more predictable events from a time series of observed variable is proposed. The procedure is applicable to time series which contains different types of events that feature significantly different predictability, or, in other words, to heteroskedastic time series. A priori selection of future events in accordance to expected uncertainty of their forecasts may be helpful for making practical decisions. The procedure first implies creation of two neural network based forecasting models, one of which is aimed at prediction of conditional mean and other - conditional dispersion, and then elaboration of the rule for future event selection into groups of more and less predictable events. The method is demonstrated and tested by the example of the computer generated time series, and then applied to the real world time series, Dow Jones Industrial Average index.

## 1. Introduction

Time series forecasting (TSF) is an important interdisciplinary branch of the modern science with numerous practical applications. The oldest and most mature approach to TSF, linear time series modeling, was developed in the framework of the mathematical theory of statistics (see, e.g., Box and Jenkins, 1976; Makridakis and Wheelwright, 1989; Priestly, 1981). Important recent developments include nonlinear deterministic technique originated from chaotic theory (e.g., Farmer and Sidorowich, 1987; Yakowitz, 1987; Sugihara and May, 1990) and neural network methodology (see, e.g., an overview by Zhang et al., 1998).

A vital characteristic of any type of a forecast is its accuracy. The general practice to estimate the accuracy of a forecast is to average definite measures of a forecast error over large enough sample of forecasts (see, e.g., Tashman, 2000) . For example, among the most popular measures of forecast uncertainty are the mean square error (MSE), the mean absolute error (MAE), the mean absolute percentage error (MAPE). Obviously, such measures do not distinguish between individual forecasts in respect with their accuracy. Meanwhile, it may be reasonable to expect that, in some cases, predictability of different events regularly differs. Such situation may take place, for example, when a time series contains the events of two (or more) types, such that the events of the first type are more strongly dependent on the past values of observed variable than the events of another type. Thus an appropriate forecasting system will be able to predict the events of the first type with greater, on the average, accuracy. When this is a case, the information concerning the accuracy of *individual* forecasts may be very useful for decision making.

This paper proposes the method aimed at a priori estimation of accuracy of a forecast of a given future event and pre-selection of future events in accordance with expected accuracy of their forecasts. In other words, we propose the method for separating more predictable events from less predictable. As a result, we are able to predict a certain fraction of events with higher, than on the average, accuracy. It is quite reasonable to expect that not only a future event itself is dependent on past events but an error of a forecast of a given event is also pre-determined, to a certain degree, by past values of the relevant variable. When forecasted values is equal to their true mathematical expectations, the mathematical expectation of a forecast error is equal to a variance of a conditional probability distribution function (CPDF) of a time series variable. Thus an appropriate forecasting system which is designed to take into account variability of forecasting uncertainty should be aimed at predicting both conditional mean and conditional variance (dispersion).

There is a lot of publications dealing with predictions of a variance of CPDF within the framework of generalized autoregressive conditional heteroskedasticity (GARCH) models (see, e.g., an overviews by Bollerslev, 1992, and Bera and Higgins, 1993). The main area of practical applications of those studies is prediction of volatility of financial and stock markets. The first GARCH models were based on the assumption of normal distribution of conditional probability. More recently, this assumption was significantly relaxed. In particular, it was suggested to approximate the shape of CPDF by means of Hermit polynomials (Gallant and Tauchen, 1989), or to use a nonparametric density estimator (Engle and Gonsales-Rivera, 1991; Boudoukh et al., 1997). However, those models are usually aimed at prediction of conditional variance which is considered (strictly speaking, not always legitimately) as a measure of volatility of an observed characteristics, while predictions of the conditional mean is, as a rule, done very roughly and are not analyzed. That is, at present, GARH models cannot be considered as an available framework useful from the point of view of the objectives specified above.

While the GARCH modeling approach assumes explicit determination of a shape of CPDF, there is a large variety of forecasting methods which are based on *ad hoc* rules for model optimization. Those methods, however, deal with direct forecasting of future events in time series, rather than estimations of any parameters of CPDF or predictions of errors of forecasts. One of the most powerful and popular forecasting tools is provided by artificial neural network methodology (see, e.g., overviews by Widrow et al., 1994, Hill et al., 1994, Zhang et al., 1998, and Gardner and Dorling, 1998). The choice of a cost function for model optimization is usually related to aims of a study. In particular, the forecasting model optimized in terms of MSE is expected to provide the best estimation of mathematical expectation of the future value.

In this paper, we use the standard neural network methodology for predicting magnitudes of errors of forecasts performed with another neural network based model. Our method combines both nonlinearity and flexibility inherent to neural networks. It is more straightforward and easy to implement if compared with nonlinear modifications of ARCH models. Perhaps, the closest prototype for our method is a method proposed by Ginzburg (1994), Ginzburg and Horn (1994). They suggested to reduce *systematic* errors of forecasts by means of a special neural network trained to predict the errors of the main model. In contrast to Ginzburg and Horn, we model the absolute values of errors rather than the errors themselves. In our case, information on sign of the errors is excessive. Furthermore, they simply add the predicted error to the forecasted variable value, while we use it as an indicator of predictability of a given future event. The idea of pre-selection of forecasted events in relation to their a priori defined property has also been exploited to some extent in literature earlier. Namely, Povinelli and Feng (1998) and Povinelli (2001) have proposed a specific time series data mining framework aimed at a priori identification of the future events with the pre-defined (desired) properties. The key idea of their method is selection of a number of patterns from the past evolution that maximize a so called event characterization function defined a priori. They applied this method to stock market time series in order to identify buying opportunities which would result in positive returns. Despite some similarity of the ideas of our methods with the ideas exploited in the earlier contributions, both the aims and technical implementation of our method are different. The full description of the method is given in Section 2.

Naturally, the practical usefulness of our method will be maximal in situations when a given time series contains actually both more predictable and less predictable events. In order to get better idea of such situations, we present in Section 3 the computer generated time series for which a forecasting error is event-dependent, and discuss application of our method to those time series. The method is then applied (Section 4) to the real world time series, Dow Jones Industrial Average stock index.

## 2. Description of the method
### 2.1. The basic ideas

Assume that we have a scalar time series, $x_1$, $x_2$, $x_3$ ... , where indexes of a variable correspond to consecutive moments of time. We can always write the following formal relationship:

$$x_{i+1} = f(x_i, x_{i-\tau 1}, x_{i-\tau 2}, \ldots, x_{i-\tau m}) + \varepsilon_{i+1}, \ i = 1 + \tau_m, 2 + \tau_m \ldots, \tag{1}$$

where $f(\bullet)$ is an arbitrary real function of m variables, $\tau_1 < \tau_2 < \ldots < \tau_m$ are time delays, and $\varepsilon_{i+1}$ is a real number. The primary assumptions of time series forecasting are that there is a regular causal relationship between past and future events (here, variable values), and that this relationship can be approximated by a function of several time-delayed variables. Correspondingly, the problem of one-step forecasting is to determine such values of m and $\tau_i$ (i=1, ..., m), and to find out such $f(\bullet)$ that a certain measure of $\varepsilon_{i+1}$, e.g., $\varepsilon_{i+1}{}^2$, averaged over a forecasting period be as small as possible. As it was noted in Introduction, there are several approaches to solving this problem. In our opinion, one of the most powerful and universal way for fitting causal relationships is based on neural network methodology.

Our method assumes building two different neural network models. First model is quite traditional. It is aimed to approximate the actual causal relationship between past and future values and provide forecasts of future events (see(1)). The second model is more specific. It is aimed to predict the absolute error of a forecast of a future event $x_{i+1}$ by utilizing a set of present and past time delayed variables. That is, the neural network is expected to approximate an unknown causal function

$$g(x_i, x_{i-\tau 1}, x_{i-\tau 2}, \ldots, x_{i-\tau m}) = |\varepsilon_{i+1}| + \delta_{i+1}, \ i = 1 + \tau_m, 2 + \tau_m, \ldots, \tag{2}$$

such that a sum of squared values of residuals $\delta_{i+1}$ for forecasting period should be as small as possible. The estimated magnitudes of errors of forecasts $|\varepsilon_{i+1}|$ are then used for classifying future events in respect to predicted errors of forecasts. The example of the rule for such classification is discussed in the next section.

### 2.2 The algorithm

Let us discuss now our method in more detail. It employs usual feedforward multilayer neural networks (multilayer perceptron) with one hidden layer. This kind of networks appears to be most popular in various applications (see, e.g., overviews by Widrow et al. (1994), Hill et al. (1994), Zhang et al. (1998), and Gardner and Dorling, (1998)). It has been proved that the perceptron with just one hidden layer is capable of approximating any measurable function to any desired degree of accuracy (Hornik et al., 1989; Hornik, 1991). Mathematically, the perceptron $p(\mathbf{X})$ with one hidden layer and logistic activation function can be represented as follows:

$$p(\mathbf{X}) = \sum_{j=1}^{N_N} \left( w_j g_j(X) + w_0 \right), \ \ g_j(X) = \frac{1}{1 + exp\left[-\sum_{i=1}^{n}(\overline{w_{ij}} \, x_i + \overline{w_{0j}})\right]}, \tag{3}$$

where $\mathbf{X} = (x_1, \ldots, x_n)$ is a vector of input values, $w_j$, $w_0$, $\overline{w_{ij}}$, $\overline{w_{0j}}$ are weight coefficients, $N_N$ is a number of neurons in the network. In our case, input values are time delayed observations (see (1) and (2)).

The key element of any neural network methodology is a procedure by which appropriate values of weights are determined. Such a procedure is usually referred to as a training algorithm. The training is achieved in the process of minimization of a certain cost function. We employ a most typical cost function which is the sum of squared errors:

$$E = \frac{1}{N_t} \sum_{i=1}^{N_t} \left( x_i - a_i \right)^2, \tag{4}$$

where $x_i$ is an observed value (a target), $a_i = p(\mathbf{X})$ is an output of the network, and $N_t$ is total number of training cases. One of the earliest and most popular training method is the backpropagation (BP) algorithm (e.g., Reed and Marks, 1999). The main drawback of this algorithm is its slow convergence. The more efficient methods are based on conjugate gradients, quasi-Newton, or Levenberg-Marquardt algorithms. However, our own experience shows that these methods often suffer from the lack of robustness and stability of convergence, especially when data are very noisy. After some experimenting, we came to the Polytope method (which is often referred to as the downhill simplex method) (see, e.g., Gill et al., 1981; Press et al., 1991). Originally, this method was suggested by Nelder and Mead (1965). In contrast to the other algorithms, this method does not employ any derivatives of the cost function. Due to this, the Polytope algorithm is very robust and never diverges. The rate of convergence is problem-dependent. From our own experience of working with very noisy data (such as financial time series), we learn that the Polytope algorithm may be as fast as conjugate gradient method and much faster than standard BP method. Though using Polytope algorithm for function optimization does not appear to be unusual (see, e.g., Carley and Morgan, 1989; Nash et al., 1987), we are aware of only two studies employing this algorithm for training neural networks. Namely, Curry and Morgan (1997) performed some experiments with neural network trained with both the Polytope and BP algorithms, and obtained better results with the former. Curry et al. (2002) have employed the Polytope algorithm along with BP for training networks aimed to model the price-quality relationships. They again found that Polytope algorithm offers superior results compared with BP. In our study, the Polytope algorithm is implemented by means of the Fortran routine presented by Press et al. (1991).

With noisy data, there is some risk that the algorithm will discover some local bad minimum of a cost function rather than a global one. The standard way to overcome the problem of bad minima is to perform multiple re-runs of the algorithm with randomly generated sets of starting values of weights, what is done in this work. On the other hand, it is well known that finding a global minimum of the cost function (4) does not guarantee that the neural network will perform well as a forecaster. When the network performs well with the training data but performs badly with unseen data, it is said that the network does not generalize, or that it is overtrained. An overtrained network tends to reproduce noise rather than regular relationships between variables. The literature suggests several ways to avoid overfitting such as model selection via bootstrapping or cross-validation, early stopping, jittering, and some others (see, e.g., Sarle, 1995; Sarle, 2002). However, none of those ways appear to be efficient in all situations. In this study, we combine early stopping (see, e.g., Nelson and Illingworth, 1991) and model selection via bootstrapping (Efron and Tibshirani, 1993). Consequently, our training algorithm is organized as follows.

1. Given $N_t$ training cases, chose randomly $N_t/2$ cases and run the Polytope optimization routine for these cases with random starting values of the weights. After each internal iteration of the polytope algorithm, calculate a value of a cost function, E (see (4)), for the remaining training cases which constitute a validation subsample of the training sample. Save that E along with values of the network weights.

2. As convergence of the Polytope algorithm is reached, select the least value among saved E values for validation sample and use corresponding weights for calculating E with all $N_t$ cases in the training sample. Save that value.

3. Repeat steps 1 and 2 for $N_1$ times. Select $N_2$ least values among $N_1$ values of E for the whole training sample. Use the corresponding sets of weights for building the forecaster as follows:

$$f = \frac{1}{N_2} \sum_{n=i_{b1}}^{i_{bN_2}} p(X, n) \,, \qquad (5)$$

where $f$ is a forecasted value, n is an index of a network, and $i_b$ are the indexes of the "the best" networks. Averaging over a large enough set of individual networks (which may be regarded as a partial case of combining neural networks) makes sense as neither early stopping nor bootstrapping applied separately or in combination do not always provide the best network from the point of view forecasting. Despite those precautions against overtraining, there is still some risk that a single network may be seriously biased toward the training sample and its forecasts will be spurious. Averaging over several networks is intended to diminish this risk and enhance stability of forecasts.

The steps 1-3 are repeated twice with two different training samples. Initially, the algorithm utilizes the training sample that consists entirely of time delayed variables. As a result, we get a standard time series forecaster. Then, this forecaster is applied again to its own training sample in order to evaluate absolute errors of the fit for each training case. Those errors are used as targets in the second training sample which includes the same predictive variables as the first training sample. As a result, we get a network model which is aimed to predict errors of forecasts performed with the first network model. To demonstrate that predicted errors of forecasts have indeed some discriminative power, the following procedure of classifying forecasts in accordance to predicted errors is used.

Let $\varepsilon_i$ (i=1,2 … $N_t$) be the errors of the fit of the first model. These errors are to be arranged orderly in respect to their magnitudes, that is, $\varepsilon_{i1} < \varepsilon_{i2} < \ldots < \varepsilon_{iNt}$. This sequence is divided into two (or more) parts such that each part contains equal number of the points. The boundary values of $\varepsilon$ determine the cells for sorting out forecasted events in accordance to an expected error of forecasts. That is, two closest boundary values, one of which is less than forecasted error, and the other is greater than that, define the cell in which that error is placed. Each set of future events with forecasted errors belonging to the same cell is considered separately. The number of cells is dependent on a concrete task In this study, we divide future events into only two sets which include more predictable and less predictable events. To ensure that all of the forecasted events are taken into account, the limiting values of $\varepsilon$, $\varepsilon_{i1}$ and $\varepsilon_{iNt}$, are shifted to zero and infinity, respectively.

### 2.3. Parameter values

The parameters which have to be set a priori include the number of time-delayed variables, m (see (1) and (2)), the time delays, $\tau_i$, the number of neurons in the networks, $N_N$, and the parameters of the network selection and combination procedure, $N_1$ and $N_2$. As far as we know, there is no universal method for selecting any of these parameters besides the trial-and error method. Usually, the parameters of a forecaster are determined with consideration for goals of a study and ideas regarding underlying mechanism of the time series generation. The primary goal of this paper is demonstration that pre-selection of future events in relation to the expected accuracy of forecasts is possible in principle. Accordingly, to insure that the positive results of this study is not just a consequence of "lucky" choice of parameters, we avoid "fine tuning" of the algorithm. Instead, we apply the same algorithm with the same parameters (the only exception is discussed below) to four different time series and obtain positive results with each of the series. No doubt that tuning the algorithm separately for each time series would allow us to get better results.

Since it was assumed in initial stage of our study that the method will be applied to financial time series, we performed some preliminary experiments with the Dow Jones Industrial Average monthly mean data and found that two different forecast methods, nearest neighbors (Farmer and Sidorowich, 1987) and neural networks give, generally, better forecasts (without the pre-selection of events) when the number of predictive variables (or, in other words, embedding dimension) equals 2. Consequently, we fix m=2 in the rest of this paper. The computer models used in this study for generating time series are also defined with two independent variables.

The time delay was defined following to methods which are commonly used in chaotic time series analysis (see, e.g., Abarbanel, 1996). These are mutual information and autocorrelation function methods. For Dow Jones Industrial Average monthly means series, both the methods indicate that $\tau=3$. The time delay between independent variables in the computer generated time series equal 1, in accordance with formulation of the models. The difference in time delays is the only difference between the algorithms applied to the computer generated and the real world time series.

We put the number of neurons $N_N=4$. Our experience shows that our neural network models give somewhat worse results with smaller $N_N$. The increase of $N_N$ increases significantly computational burden but does not change the results considerably. For purely computational considerations, we set the number of trials $N_1=50$. With a large enough value of $N_1$, the results are dependent, mostly, on ratio of $N_1$ to $N_2$ rather than $N_1$ and $N_2$ separately. After some experimenting we set the number of combined networks $N_2=25$. Note that both the forecaster of events and forecaster of errors have the same parameter values.

Finally, let us discuss the specification of training sample and test sample. At present, the literature does not offer any universal method for specifying an optimal size of training and test samples for a given problem and a data set. However, we have to take account several requirements and limitations, the part of which is dictated by specific tasks of our study. (1) It is generally understood that both training and testing samples should be sufficiently large in order to be representative of the data and the underlying mechanism of their generation. (2) As it was noted above, the main idea concerning organization of this study is that all data sets should be processed in the same way. Therefore, the total size of the computer generated time series can not exceed the size of the real world time series which has about 1200 data points. (3) To obtain more statistically sound conclusions, it is desirable to consider several independent cases with each of the time series. In order to meet these requirements and limitations, the time series are processed in the following way.

Starting from the initial point of a given time series, we assign 200 consecutive points to the training sample. The next 100 points are assigned to the test sample. After the neural network is built and results are obtained, both the training sample and test sample are shifted 100 points ahead and all the procedure is repeated. Clearly, that in each iteration, forecasts are made with entirely different test samples. Therefore, with a time series which consists of 1200 point, we get 10 independent examples of application of our method.

### 3. Application of the method to the computer-generated time series

#### 3.1. Description of the computer models for time series generation

_Model I:_ The random chain with non-uniform predictability of transitions

Let us assume that a set of possible values of model variable consists of only two numbers, -1 and 1. Assume further that each variable value $x_{i+1}$ ($i=1,2,\ldots$) may depend only on two closest past variable values, $x_i$ and $x_{i-1}$, with conditional probabilities of transitions defined as follows:

$$p(x_{i+1}=1 \mid x_i=1; x_{i-1}=1)=0.2; \quad p(x_{i+1}=-1 \mid x_i=1; x_{i-1}=1)=0.8;$$
$$p(x_{i+1}=1 \mid x_i=-1; x_{i-1}=-1)=0.2; \quad p(x_{i+1}=-1 \mid x_i=-1; x_{i-1}=-1)=0.8; \qquad (6)$$
$$p(x_{i+1}=1 \mid x_i=-1; x_{i-1}=1)=0.5; \quad p(x_{i+1}=-1 \mid x_i=-1; x_{i-1}=1)=0.5;$$
$$p(x_{i+1}=1 \mid x_i=1; x_{i-1}=-1)=0.5; \quad p(x_{i+1}=-1 \mid x_i=1; x_{i-1}=-1)=0.5.$$

It is clear that when signs of $x_i$ and $x_{i-1}$ are different, two possible predictions of $x_{i+1}$ are equal in statistical sense, and there is no regular way to predict proper values $x_{i+1}$ more frequently than in 50 cases out of 100. However, when $x_{i-1}$ and $x_{i-2}$ are of the same sign, the prediction that $x_{i+1}$ will have an opposite sign to that of $x_i$ and $x_{i-1}$ is obviously preferable. As a result, approximately 40% of the events generated in accordance to the rule (6) are more predictable and 60% are less predictable (or, better to say, unpredictable at all).

Since the neural networks which are employed in this study are not suited to work with discrete numbers, we "dressed" the chain with moderate normally distributed "measurement" noise. Specifically, each variable value $x_i$ in the chain was replaced by a value $x_i+v$, where v was sampled from the normal distribution with the zero mean and $\sigma=0.3$. Though introducing the additional noise worsens the absolute predictability of the time series, it is still useful, since it allows us to build more "robust" neural network models, and, as a consequence, to obtain more uniform results.

*Model II:* The time series with mixed data types

The model illustrates a situation when data are generated by a complex enough system which has two (or more) qualitatively different regimes of behavior. The system changes regimes frequently (on time scales of the whole time series) in response to variations of some external factors. Our model system possesses two qualitatively different regimes.

The first regime is deterministic chaos defined by the well known chaotic system which is the Henon map (Henon, 1979)

$$x_{i+1}=1 - \alpha\, x_i^2 +\beta\, x_{i-1}\,,\ \ i=1,2\dots \tag{7}$$

with parameters $\alpha=1.4$ and $\beta=0.3$. In spite of the fact that time series generated with (7) is chaotic, it is obvious that the given values of $x_{i-1}$ and $x_i$ determine the "future" value of x, $x_{i+1}$, exactly. That is, while a one-step forecast is concerned, the time series (7) is totally predictable. The unpredictability inherent to chaotic systems arise due to loss of information about initial conditions after multiple iterations (see, e.g., Moon, 1987). The horizon of predictability is dependent on largest Lyapunov exponent and desired accuracy of predictions.

The second regime of our model system is generation of white noise: The data are taken from the normal distribution $N(0, \sigma_H)$, where $\sigma_H$ is the standard deviation of chaotic data generated by the Henon map (7). This regime is, obviously, totally unpredictable.

The model changes regimes randomly, such that duration of the chaotic deterministic regime ranges from 1 to 58 steps and duration of noisy regime ranges from 1 to 49 steps. With such defined duration of regimes, the time series generated by our system contains an approximately equal number of predictable and unpredictable events. Note that in order to be predictable, the given data point in our time series should be preceded by at least two points generated by the Henon map. Note also that the noisy parts of the time series may contain the same pairs of variable values ($x_{i-1}$, $x_i$) as the chaotic parts, so that the neural networks is not able to distinguish between two types of data quite definitely. However, while the noisy data cover the plane ($x_{i-1}$, $x_i$) quasi-continuously, the chaotic points are located in this plane in an extremely non-uniform manner in accordance with the fact that fractal dimension of the chaotic attractor of the Henon map is rather close to unity (about 1.2) (Grassberger and Procaccia, 1983). This observation means that certain areas of the plane ($x_{i-1}$, $x_i$) contain a much larger, than on the average, fraction of "chaotic" points and, thus, the chaotic data indeed keep higher predictability in our time series.

*Model III:* The random chain with uniform predictability of transitions

This model is designed to provide a counter example for the above two cases. It is quite similar to model 1, except that all possible *predictive* states (that is, pairs ($x_i$, $x_{i-1}$) in this case) are equivalent from the point of view of predictability of the next event. The conditional probabilities of transitions are defined as follows:

$$\begin{aligned}
&p(x_{i+1}=1\mid x_i=1;\ x_{i-1}=1)=0.2; \quad p(x_{i+1}=-1\mid x_i=1;\ x_{i-1}=1)=0.8;\\
&p(x_{i+1}=-1\mid x_i=-1;\ x_{i-1}=-1)=0.2; \quad p(x_{i+1}=1\mid x_i=-1;\ x_{i-1}=-1)=0.8;\\
&p(x_{i+1}=1\mid x_i=-1;\ x_{i-1}=1)=0.2; \quad p(x_{i+1}=-1\mid x_i=-1;\ x_{i-1}=1)=0.8;\\
&p(x_{i+1}=-1\mid x_i=1;\ x_{i-1}=-1)=0.2; \quad p(x_{i+1}=1\mid x_i=1;\ x_{i-1}=-1)=0.8.
\end{aligned} \tag{8}$$

The chain generated with these probabilities contains an equal, asymptotically, numbers of the four possible predictive states. For each of those states, we have equal chances to predict the next value properly. To process all the time series in consistent manner and to show more clearly that when the actual difference in predictability is absent, the forecaster does not discriminate events, a half of the possible transitions in the time series were marked arbitrarily as more predictable, and a half as less predictable ones.

Before concluding this section, let us mention one more possible typical situation with non-uniform predictability. Imagine chaotic dynamic system with distinctly non-uniform probability distribution function, or, in other words, with non-uniform density of the chaotic attractor. In such situation (which is rather typical, in our opinion), the time series will contain more information on some states and less information on others. Taking into account that the real time series are typically limited in length and contaminated with noise, it is reasonable to expect that more abundant states will be associated with higher accuracy of forecasts than less abundant ones. Unfortunately, this case is much more difficult to specify and analyze than the cases defined above. Due to this, its consideration falls outside the scope of this paper.

### 3.2. Results

Tables 1-3 show the results obtained with the time series generated by the models I-III, respectively. In particular, the tables present the RMSEs for better predictable ($\varepsilon_1$) and worse predictable ($\varepsilon_2$) sets of events selected by our forecaster, as well as for the whole test sample ($\varepsilon_t$). For convenience, all the errors are normalized over the root squared dispersion of the respective (whole) test series. The tables present also the fraction of *actually* best predictable events (as defined in the model descriptions) among the *presumably* better ($f_1$) and worse ($f_2$) predictable events selected by the forecaster, as well as among all points of the given test sample ($f_t$). The numbers of the better ($n_1$) and worse ($n_2$) predictable events identified by the forecaster are also reported. Finally, the two last columns of the tables show the correlation coefficients between actual absolute errors of the fit (or, forecast) and fitted (or, forecasted) absolute errors for the training sample ($\rho_t$) and test sample ($\rho_p$). The bottom line of the tables reports the corresponding average values and squared root of dispersions.

The results bear solid evidences that our forecasting models are able to distinguish, though not absolutely perfectly, the more predictable events from less predictable ones, if such difference actually exists. Indeed, with the time series generated by the Model I, a set of better predictable events selected by the forecaster has lower RMSE than a whole test set at least in 8 cases out of 10. In all cases but one a fraction of actually better predictable events in the selected set of presumably better predictable events is much larger than both in the set of presumably worse predictable events and the whole test set. The independent means statistical test shows that the hypothesis of equivalence of corresponding average values should be rejected with statistical significance level less than 5% (in other words, the probability of Type I error is less than 5%).

With the time series generated by the Model II, a set of better predictable events selected by the forecaster has lower RMSE than a whole test set in all 10 cases. Similarly, in all cases a fraction of actually better predictable events in a selected set of a (presumably) better predictable events is significantly higher than in the whole test set. The independent means statistical test bears evidence that means of $\varepsilon_t, \varepsilon_1$, and $\varepsilon_2$ are different with probability of Type 1 error less than 5%. The fractions $f_1$ and $f_2$ defined above are different with probability of Type 1 error less than 5%. The fractions $f_1$ and $f_t$ are different with probability of Type 1 error less than 10%.

As for results obtained with the time series generated by the Model III, they are also quite expected. Indeed, the forecaster failed to select a set with less than the mean forecast error in 6 cases from 10. The average $\varepsilon_1$ is even somewhat larger than the average $\varepsilon_t$. Average fractions $f_1$, $f_2$, and $f_t$ are identical, as expected. Note also that the mean of the correlation coefficient ($\rho_p$) between actual absolute errors of the forecast and forecasted absolute errors for the training sample is much less than those obtained for the Models I and II and very close to zero. The last observation means that the neural network is not able to recognize any regular relationship between variable values and forecast error in this case.

## 4. Application of the method to the Dow Jones Industrial monthly average time series

Forecasting of stock indexes has always gained a lot of attention of both practitioners and scientists. It was long ago established that the stock prices time series are extremely noisy and difficult to forecast. The theoretical basis for this empirical fact provides the "weak form" of the "efficient market hypothesis" (Fama, 1965; Malkiel, 1985; Fama, 1991), in accordance to which price changes are independent on the past. Nonetheless, there are growing evidences that financial time series, including stock indexes, are actually predictable to some degree. There is no place to review a voluminous literature on this subject here. As examples of contributions in favor of predictability and long "memory" of market prices, let us note only the studies by Lo and MacKinlay (1988), Sornette et al. (1996), and Grau-Carles (2000). Our results which follow are in agreement with the notion of weak determinism of stock prices.

The data that we deal with in this section are Dow Jones (DJ) Industrial Average (close) monthly means from January 1900 to December 2000. The data were obtained from the Internet site Economagic.com: Economic Time Series Page (http://www.economagic.com/). Instead of modeling the raw data, we consider their normalized differences:

$$x_i = 2 \frac{y_i - y_{i-1}}{y_i + y_{i-1}} \qquad (9)$$

where $y_i$ and $y_{i-1}$ are consecutive data points of the original time series, and $x_i$ is a variable of the converted time series. The transformation (9) is aimed at reducing non-stationarity of the original time series.

The result obtained after application of our method to DJ time series are reported in Table 4. Just as Tables 1-3, Table 4 shows normalized RMSE ($\varepsilon_1$, $\varepsilon_2$, $\varepsilon_t$) of forecasts for different sets of forecasted events, the numbers of better predictable and worse predictable events ($n_l$, $n_2$), and correlation coefficients, $\rho_t$ and $\rho_p$ defined above (see Section 3). Besides, the table presents two additional statistics which are U statistics ($u_l$, $u_2$, $u_t$), and the fractions of properly forecasted signs of changes of DJ index. The U statistics is the ratio of the RMSE of a model's forecasts to the RMSE of the random walk forecasts. The random walk forecast of the next month's stock price is equal to the current stock price. That is, the random walk predicts $x_{t+1} = 0$. While the U statistic is less than 1, a forecasting model outperforms the random walk model, and vice versa. Note that while RMSEs $\varepsilon_1$, $\varepsilon_2$, $\varepsilon_t$ are normalized over the squared root of dispersion of of the whole test sample, the RMSEs for the random walk were evaluated for each of the subsets of the training set separately. Predicting direction of the future price change rather than the value of relevant variable may be of some interest from the practical point of view. If the price changes followed the random walk, the numbers of positive and negative price changes would be equal. However, the DJ behavior demonstrates a well pronounced positive trend, such that the yearly average value of DJ index in the beginning of the twentieth century was more than 100 times less than that in the end of the century. Johansen and Sornette (2001) argue that this trend cannot be adequately described by neither linear, nor exponential, nor power laws, and present a more complex law. Due to this trend, positive changes of monthly DJ constitute 58.6% in the twentieth century time series.

The results indicate that our method is indeed capable to distinguish more predictable events from less predictable ones in the DJ time series. In particular, all the statistics evaluated for the sets of more (as expected) predictable events are better than the statistics evaluated for the sets of less predictable events in 7 cases out of 10. The difference between the selected sets is somewhat more significant if the statistics are considered separately. Specifically, $\varepsilon_1$ ($U_1$) is less than $\varepsilon_2$ ($U_2$) in 8 cases out of 10, and $S_l$ is greater than $S_2$ again in 8 cases. The independent means statistical test bears evidence that both means of $\varepsilon_1$ and $\varepsilon_2$, and $\varepsilon_1$ and $\varepsilon_t$ are different with probability of Type 1 error less than 5%. (The corresponding t-values are 3.28 and 2.16, while the critical t-value in accordance to the Student's distribution for significance level $\alpha=0.05$ is 2.01). The same conclusions are kept for U statistics and S statistics. (The t-values are 3.51 and 2.12; 3.53 and 2.23, respectively.)

Note that though the investigation of sensitivity of the results to the parameters of our algorithm is beyond the scope of this paper, we repeat all the experiments described in this paper with different size of the training set. Namely, in addition to the main case with the number of point in the training set $N_t=200$, we performed the same experiments with $N_t=100$ and $N_t=300$. We found that the results with $N_t=300$ are significantly better than the results with $N_t=200$, and the results with $N_t=100$ are worse than those. Specifically, in case of DJ time series and $N_t=300$, the selection of more and less predictable events was successful with regard to all of the statistics in all (nine) experiments.

## 5. Conclusions

This paper focuses on the idea of a priori classification of the future events in relation to an expected accuracy of their forecasts. We believe that such classification may be very useful for decision making and for reveling the nature of underlying data generation mechanism. We propose the neural network based method for a priori assessing an error of a forecast of a given event and discriminating between more and less predictable events. The main idea of the method is creation of the special neural network model aimed to predict absolute errors of forecasts. The discriminating power of the method is demonstrated first by the example of the computer generated time series. It is shown that when predictability of different events in a time series differs, the method is able to select such sets of future events that contain, for the most part, actually more predictable, and less predictable events, in accordance to formulation of a data generating model. The method is then successfully applied to the real world time series, Dow Jones Industrial Average monthly means. The results of a priori classification of future events are given in terms of various statistics (RMSEs, U-statistics, and the numbers of forecasts with properly predicted signs of changes of the relevant variable). It is found that these measures are statistically different for the sets that should contain more predictable and less predictable events.

Note that this paper has no purpose to demonstrate maximum potential of our method. In particular, we consciously avoid fine tuning of the forecasting models for different time series considered here, in order to emphasize flexibility of our algorithm. There are obvious opportunities to enhance the discriminating power of our method and to obtain more accurate forecasts for more predictable events. First, all the parameters of the algorithm should be carefully adjusted in case of each of given time series. Second, it may be advantageous to increase the number of cells for pre-selecting future events. This may be especially useful when the time series is expected to contain events of more than two different types. Third, since the main forecast model is built for all cases at once, it seems to be very probable that the forecasts for more predictable events are affected by the presence of less predictable events in the time series, and vice versa. Thus, in order to improve accuracy of predictions of the more predictable events, it may be useful to build a special neural network model for the set of more predictable events once they have been identified. Finally, when a time series is noisy, it may be advantageous to employ one of the noise filtering methods (see, for example, Hegger and Kantz, 1998). The expanded discussion of any of this issues in this paper is well beyond the scope of this article due to space limitations. Further developments and practical applications of our method will be discussed in future papers.

*Acknowledgements.* The author is grateful to Dr. A. M. Feigin for useful discussions. The work was supported in part by the Russian Foundation for Basic Researches, grant No. 02-02-17080

Table 1.
Experimental results for the Model I[a]

| Case No. | $\varepsilon_1$ | $\varepsilon_2$ | $\varepsilon_t$ | $f_1$ | $f_2$ | $f_t$ | $n_1$ | $n_2$ | $\rho_{tr}$ | $\rho_{pr}$ |
|---|---|---|---|---|---|---|---|---|---|---|
| 1 | 0.74 | 1.04 | 0.91 | 0.83 | 0.00 | 0.39 | 47 | 53 | 0.29 | 0.45 |
| 2 | 0.79 | 0.94 | 0.86 | 0.67 | 0.00 | 0.34 | 51 | 49 | 0.44 | 0.42 |
| 3 | 1.02 | 1.01 | 1.01 | 0.77 | 0.00 | 0.41 | 53 | 47 | 0.49 | 0.19 |
| 4 | 0.84 | 1.00 | 0.92 | 0.80 | 0.00 | 0.45 | 56 | 44 | 0.43 | 0.51 |
| 5 | 0.85 | 1.07 | 1.00 | 0.97 | 0.00 | 0.35 | 36 | 64 | 0.28 | 0.36 |
| 6 | 0.91 | 1.00 | 0.96 | 0.67 | 0.00 | 0.37 | 55 | 45 | 0.49 | 0.19 |
| 7 | 0.99 | 1.00 | 0.99 | 0.75 | 0.02 | 0.45 | 59 | 41 | 0.37 | 0.23 |
| 8 | 0.90 | 1.00 | 0.96 | 0.33 | 0.40 | 0.37 | 48 | 52 | 0.23 | 0.23 |
| 9 | 0.91 | 1.05 | 1.00 | 0.79 | 0.04 | 0.36 | 43 | 57 | 0.36 | 0.39 |
| 10 | 0.85 | 1.00 | 0.93 | 0.74 | 0.04 | 0.39 | 50 | 50 | 0.42 | 0.42 |
| Mean/ Devi- ation | **0.88/ 0.095** | **1.01/ 0.036** | **0.95/ 0.049** | **0.73/ 0.17** | **0.05/ 0.12** | **0.39/ 0.039** | **49.8/ 6.7** | **50.2/ 6.7** | **0.38/ 0.090** | **0.34/ 0.12** |

[a]The table reports the RMSEs for better predictable ($\varepsilon_1$) and worse predictable ($\varepsilon_2$) sets of events selected by our forecaster, and for the whole test sample ($\varepsilon_t$). All RMSEs are normalized over the root squared dispersion of the respective (whole) test series. The table present also the fraction of actually better predictable events (as defined in the model descriptions) among the presumably better ($f_1$) and worse ($f_2$) predictable events (as selected by the forecaster), as well as among all points of a given test sample ($f_t$); the numbers of the better ($n_1$) and worse ($n_2$) predictable events identified by the forecaster; the correlation coefficients between actual absolute errors of the fit (or, forecast) and fitted (or, forecasted) absolute errors for the training sample ($\rho_t$) and test sample ($\rho_p$). The bottom line reports the corresponding average values and root squared dispersions.

Table 2.
Experimental results for the Model II[a]

| Case No. | $\varepsilon_1$ | $\varepsilon_2$ | $\varepsilon_t$ | $f_1$ | $f_2$ | $f_t$ | $n_1$ | $n_2$ | $\rho_{tr}$ | $\rho_{pr}$ |
|---|---|---|---|---|---|---|---|---|---|---|
| 1 | 0.90 | 1.02 | 0.96 | 0.67 | 0.35 | 0.51 | 49 | 51 | 0.41 | 0.07 |
| 2 | 0.74 | 1.09 | 0.95 | 0.70 | 0.36 | 0.51 | 0.44 | 56 | 0.50 | 0.42 |
| 3 | 0.80 | 0.97 | 0.90 | 0.75 | 0.39 | 0.55 | 44 | 56 | 0.38 | 0.10 |
| 4 | 0.67 | 0.96 | 0.80 | 0.93 | 0.68 | 0.83 | 60 | 40 | 0.32 | 0.06 |
| 5 | 0.58 | 1.10 | 0.88 | 0.72 | 0.58 | 0.65 | 50 | 50 | 0.58 | 0.38 |
| 6 | 0.89 | 1.06 | 1.00 | 0.44 | 0.23 | 0.31 | 39 | 61 | 0.53 | 0.14 |
| 7 | 0.83 | 1.06 | 0.94 | 0.79 | 0.63 | 0.72 | 57 | 43 | 0.24 | 0.16 |
| 8 | 0.82 | 0.92 | 0.87 | 0.63 | 0.5 | 0.57 | 52 | 48 | 0.26 | 0.05 |
| 9 | 0.83 | 0.87 | 0.85 | 0.56 | 0.47 | 0.51 | 43 | 57 | 0.28 | 0.08 |
| 10 | 0.70 | 0.92 | 0.80 | 0.81 | 0.47 | 0.66 | 57 | 43 | 0.38 | 0.24 |
| Mean/ Devi-ation | **0.78/ 0.10** | **1.00/ 0.080** | **0.90/ 0.068** | **0.70/ 0.14** | **0.47/ 0.14** | **0.58/ 0.14** | **49.8/ 6.7** | **50.2/ 6.7** | **0.39/ 0.12** | **0.17/ 0.13** |

[a]The table reports the same statistics as the table 1.

Table 3.
Experimental results for the Model III[a]

| Case No. | $\varepsilon_1$ | $\varepsilon_2$ | $\varepsilon_t$ | $f_1$ | $f_2$ | $f_t$ | $n_1$ | $n_2$ | $\rho_{tr}$ | $\rho_{pr}$ |
|---|---|---|---|---|---|---|---|---|---|---|
| 1 | 0.86 | 0.73 | 0.80 | 0.52 | 0.50 | 0.51 | 52 | 48 | 0.27 | 0.1 |
| 2 | 0.87 | 0.70 | 0.80 | 0.29 | 0.67 | 0.45 | 58 | 42 | 0.35 | 0.05 |
| 3 | 0.97 | 0.79 | 0.90 | 0.72 | 0.28 | 0.54 | 60 | 40 | 0.42 | -0.08 |
| 4 | 0.82 | 0.77 | 0.80 | 0.56 | 0.50 | 0.53 | 54 | 46 | 0.28 | -0.03 |
| 5 | 0.78 | 0.79 | 0.79 | 0.47 | 0.53 | 0.50 | 55 | 45 | 0.37 | 0.14 |
| 6 | 0.79 | 0.85 | 0.82 | 0.49 | 0.50 | 0.49 | 46 | 54 | 0.12 | 0.05 |
| 7 | 0.70 | 0.81 | 0.76 | 0.48 | 0.44 | 0.46 | 52 | 48 | 0.10 | 0.20 |
| 8 | 0.75 | 0.86 | 0.81 | 0.47 | 0.55 | 0.51 | 47 | 53 | 0.29 | 0.16 |
| 9 | 1.04 | 0.85 | 0.97 | 0.41 | 0.5 | 0.45 | 56 | 44 | 0.34 | -0.18 |
| 10 | 0.93 | 0.93 | 0.93 | 0.56 | 0.54 | 0.55 | 50 | 50 | 0.16 | 0.17 |
| Mean/ Devi-ation | **0.85/ 0.11** | **0.81/ 0.067** | **0.84/ 0.069** | **0.50/ 0.11** | **0.50/ 0.10** | **0.50/ 0.036** | **53.0/ 4.5** | **47.0/ 4.5** | **0.27/ 0.11** | **0.06/ 0.12** |

[a] The table reports the same statiscics as the table 1.

Table 4.
Experimental results for the Dow Jones Industrial Average monthly means time series[a]

| Case No. | $\varepsilon_1$ | $\varepsilon_2$ | $\varepsilon_t$ | $U_1$ | $U_2$ | $U_t$ | $S_1$ | $S_2$ | $S_t$ | $n_1$ | $n_2$ | $\rho_{tr}$ | $\rho_{pr}$ |
|---|---|---|---|---|---|---|---|---|---|---|---|---|---|
| 1 | 0.98 | 0.93 | 0.95 | 0.99 | 0.92 | 0.95 | 0.61 | 0.72 | 0.67 | 46 | 54 | 0.20 | 0.04 |
| 2 | 0.56 | 1.29 | 0.98 | 0.87 | 1.00 | 0.97 | 0.75 | 0.55 | 0.66 | 51 | 49 | 0.39 | 0.43 |
| 3 | 0.86 | 1.03 | 0.96 | 0.99 | 0.94 | 0.96 | 0.58 | 0.58 | 0.58 | 43 | 57 | 0.46 | 0.07 |
| 4 | 0.99 | 1.12 | 1.02 | 0.93 | 1.12 | 0.99 | 0.68 | 0.55 | 0.64 | 76 | 24 | 0.42 | 0.08 |
| 5 | 0.96 | 1.03 | 0.99 | 0.88 | 1.03 | 0.94 | 0.74 | 0.36 | 0.58 | 58 | 42 | 0.23 | 0.13 |
| 6 | 0.83 | 1.09 | 0.99 | 0.95 | 0.99 | 0.98 | 0.67 | 0.60 | 0.63 | 45 | 55 | 0.26 | 0.19 |
| 7 | 0.88 | 1.06 | 0.99 | 0.95 | 1.02 | 0.99 | 0.58 | 0.38 | 0.47 | 43 | 57 | 0.26 | 0.05 |
| 8 | 0.96 | 0.96 | 0.96 | 0.93 | 0.98 | 0.95 | 0.6 | 0.53 | 0.57 | 53 | 47 | 0.29 | 0.05 |
| 9 | 1.00 | 1.03 | 1.01 | 0.92 | 1.08 | 0.97 | 0.70 | 0.50 | 0.63 | 64 | 36 | 0.17 | 0.05 |
| 10 | 0.91 | 1.11 | 0.99 | 0.87 | 1.01 | 0.93 | 0.66 | 0.52 | 0.61 | 62 | 38 | 0.23 | 0.16 |
| **Mean /Devi ation** | **0.89/ 0.13** | **1.06/ 0.10** | **0.98/ 0.02** | **0.93/ 0.04** | **1.01/ 0.06** | **0.96/ 0.02** | **0.66/ 0.06** | **0.53/ 0.10** | **0.60/ 0.06** | **54.1/ 10.8** | **45.9/ 10.8** | **0.29/ 0.10** | **0.13/ 0.12** |

[a]See the explanatory footnote to Table 1 for definitions of $\varepsilon_1, \varepsilon_2, \varepsilon_t, n_1, n_2, \rho_{tr}$ and $\rho_{pr}$. $U_1$, $U_2$, and $U_t$ are values of U-statistics for corresponding sets of events, and $S_1$, $S_2$, and $S_3$ are fractions of forecasted events with properly predicted directions of changes of the DJ index.